\documentclass{article}
\usepackage[table]{xcolor}
\usepackage[final]{corl_2026} 

\usepackage{algorithm}
\usepackage{siunitx}
\usepackage{amsmath,amssymb}
\usepackage{cleveref}
\usepackage{booktabs}
\usepackage{multirow}
\usepackage{makecell}
\usepackage{graphicx}
\usepackage{algpseudocode}
\usepackage{bm}
\usepackage{caption}

\title{
Bridging Performance and Generalization in Reinforcement Learning for Agile Flight
}

\author{
\textbf{Jonathan Green} \hspace{0.7cm}
\textbf{Jiaxu Xing} \hspace{0.7cm}
\textbf{Nico Messikommer} \hspace{0.7cm}
\\[0.9em] \hspace{0.7cm}
\textbf{Angel Romero} \hspace{0.7cm}
\textbf{Davide Scaramuzza}\\[1.2em]
Robotics and Perception Group,
University of Zurich
\vspace{-0.5cm}
}

\begin{document}
\makeatletter
\g@addto@macro\@maketitle{
  \captionsetup{type=figure}\setcounter{figure}{0}
  \def\mycolspace{1.9mm}
  \centering
  \includegraphics[width=0.96\linewidth]
  {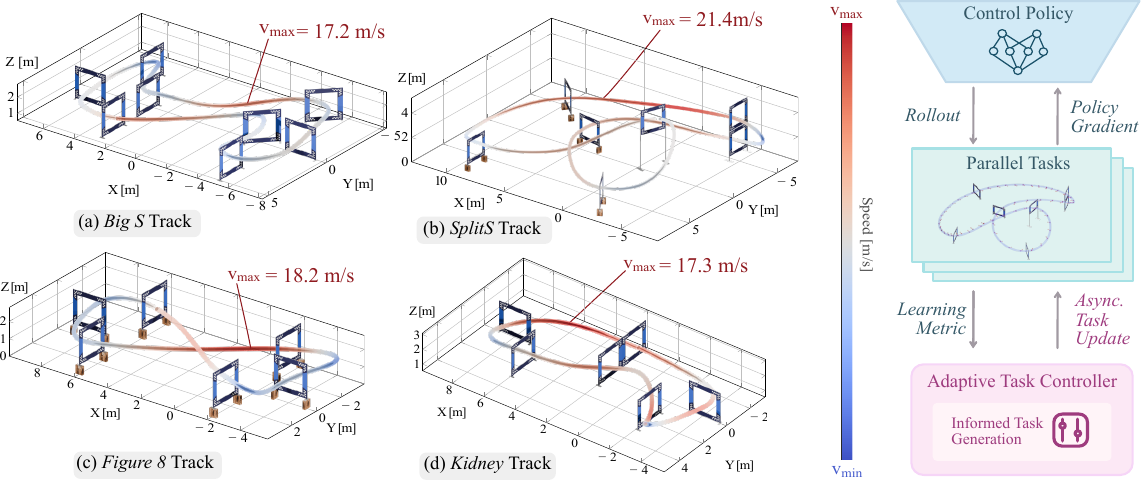}
  \vspace{0.4cm}
	\captionof{figure}{Our policy achieves zero-shot generalization on diverse unseen racetracks while maintaining performance. Trajectories shown are from real-world deployment of a single generalist policy, compared to the conventional approach of training a dedicated policy for each racetrack. We train this policy on a varied set of training tasks in parallel, asynchronously generating new tasks to replace those that are no longer useful.}
    \vspace{-0.2cm}
\label{fig:catcheye}
}
\makeatother

\maketitle

\begin{abstract}
Autonomous drone racing is a fundamentally challenging regime for autonomous aerial robots, requiring time-optimal control while operating under persistent actuation saturation.
While reinforcement learning (RL) has achieved human-level performance in this domain, current methods fail to generalize: policies trained on specific environments often crash immediately in unseen configurations. 
This failure reflects the intrinsic difficulty of zero-shot generalization in agile flight, arising from high-dimensional task variation and the tight coupling between safety and performance at high speeds.
Existing approaches that improve generalization impose a substantial cost on flight speed: control policies must significantly degrade performance to achieve even modest levels of generalization.
In this work, we propose a framework for zero-shot generalization in agile flight for RL-based drone racing.
By combining task-aware switching based on learning progress with a physically informed procedural track generator, the framework produces a fast and robust generalist policy without test-time adaptation.
Our method achieves strong zero-shot performance across a wide range of unseen racetracks in the real world, demonstrating a 7.4$\times$ improvement in generalization over the state-of-the-art approaches, while maintaining competitive racing speeds.
We validate our method's results in both simulation and real-world settings, including a challenging vision-based, end-to-end control setting that operates without explicit state estimation, where all prior approaches fail to generalize.
\end{abstract}


\section{Introduction}
Agile flight represents one of the most demanding regimes for autonomous aerial robots. 
Operating at high speeds and under aggressive maneuvers requires tight coupling between perception and control, leaving little margin for error.
Autonomous drone racing exemplifies an extreme form of agile flight, requiring sustained high-speed control and precise maneuvering through tightly constrained race tracks.
In this setting, reinforcement learning-based controllers have demonstrated the ability to achieve superhuman performance on fixed racetracks~\cite{kaufmann_champion-level_2023}.
Compared to traditional model-based control approaches, reinforcement learning can directly learn high-performance control policies through interaction with a simulated environment, using task-oriented, even non-differentiable, reward formulations~\cite{song_reaching_2023}, and naturally accommodates flexible observation modalities, such as visual inputs.
However, existing RL-based racing agents exhibit a critical limitation: they fail to generalize.
Controllers that excel on the racetracks they were trained on often crash immediately when deployed to new racetracks, typically failing to complete even a single lap~\cite{wang2025environment}. 
This behavior stands in stark contrast to human pilots, both experts and non-experts alike, who can usually complete an unseen racetrack without crashing, even on their first attempt.

Given the speeds involved in drone racing, which can be in excess of \SI{80}{\kilo\meter\per\hour} \cite{hanover_autonomous_2024}, a single crash often completely destroys the drone.
The ability to fly safely and effectively without track-specific training or fine-tuning, which we refer to as zero-shot generalization (ZSG), is therefore a fundamental capability for real-world deployment.
Prior work has taken steps toward improving ZSG for autonomous drone racing.
Wang et al. \cite{wang2025environment} demonstrated an RL control pipeline that adaptively adjusts difficulty during training by iteratively updating the layout of the training racetracks.
While this approach improved robustness on unseen race tracks, it suffered from a major drawback: the learned agents could not fly competitively fast, trading away performance for generalization.

In this work, we take a more systematic look at the ZSG problem in RL for drone racing by proposing and verifying a variety of strategies to improve ZSG.
Our framework combines large-scale parallelization with adaptive task switching and informed task generation, enabling agents to continually encounter novel challenges while avoiding overfitting. 
We characterize the factors contributing to generalization while also assessing the roles of regularization and memory. 
Our approach yields strong generalization across a wide range of unseen racetracks without sacrificing high-speed performance, achieving generalization 7.4$\times$ better than state-of-the-art methods while being 37.73\% faster. 
Our learned RL controller outperforms state-of-the-art model-based racing controllers, while retaining zero-shot deployment capability.
Furthermore, we successfully deployed it in the real world to confirm its efficacy.
To evaluate generalization under more realistic sensing constraints, we additionally consider a vision-based racing setting, in which the agent must learn directly from pixel inputs and minimal state information, a regime in which existing approaches fail to generalize.
Our method achieves successful zero-shot deployment on previously unseen tracks, demonstrating for the first time that RL-based vision policies can generalize for agile, high-speed drone racing.

Our results show that reinforcement learning can achieve strong zero-shot generalization in high-speed drone racing without a fundamental performance trade-off.
This points toward a practical path for robust, high-performance deployment of agile autonomous drones in the real world.
\section{Related Work}
\paragraph{Learning-based Agile Flight.}
The application of machine learning to agile aerial robotics has witnessed substantial progress.
Recently, autonomous drone racing has emerged as a compelling benchmark for evaluating agile flight capabilities, demanding tight integration of perception, planning, and control under stringent time constraints~\cite{hanover_autonomous_2024}.
In this domain, reinforcement learning has proven remarkably effective. Hwangbo et al.~\cite{hwangbo2017control} showed that learned policies can robustly track waypoints, while Song et al.~\cite{song_reaching_2023} demonstrated that learned controllers can surpass optimization-based methods~\cite{romero_model_2022}, attributed to their capacity for handling highly nonlinear dynamics and multi-objective optimization that challenge conventional model predictive control formulations~\cite{falanga2018pampc, xing2023autonomous}. 
This line of research culminated in Kaufmann et al.~\cite{kaufmann_champion-level_2023}, which achieved super-human performance.
Recent works have pushed toward end-to-end visuomotor control~\cite{geles_demonstrating_2024, xing2024bootstrapping}, where the networks directly learns to map visual information directly to control commands.
Beyond racing, learning-based methods have addressed safety-critical behaviors such as obstacle avoidance~\cite{zhao2024learning, zhang2024back}, further broadening real-world applicability.
\paragraph{Generalization in Reinforcement Learning.}
Significant work has explored methods to allow reinforcement learning-based agents to generalize better. 
The simplest implementation of this is Domain Randomization \cite{tobin_domain_2017}, in which a fixed set of parameters is varied randomly during training. 
More complex algorithms, such as ACCEL \cite{parker-holder_evolving_2022} and PLR \cite{jiang_prioritized_2021}, instead maintain a buffer of useful tasks, replaying some when necessary. PAIRED \cite{dennis_emergent_2021} and ALP-GMM \cite{portelas_teacher_2020} approach this topic by instead generating novel and suitably difficult tasks, using a regret-based metric and a Gaussian mixture model respectively.
In the broader robotic reinforcement learning domain, methods for improving generalization are varied. Lee et al.~\cite{lee_learning_2020} address low-dimensional task variation by producing simulation parameters with a particle filter.
For drones, Ferede et al.~\cite{ferede_end--end_2024} demonstrate a network is trained to find optimal flight trajectories, while in cluttered, obstacle-filled environments, Yu et al.~\cite{yu_mastering_2025} propose a two-stage method with increasing collision enforcement.
\begin{figure*}
    \centering
    \includegraphics[width=\linewidth]{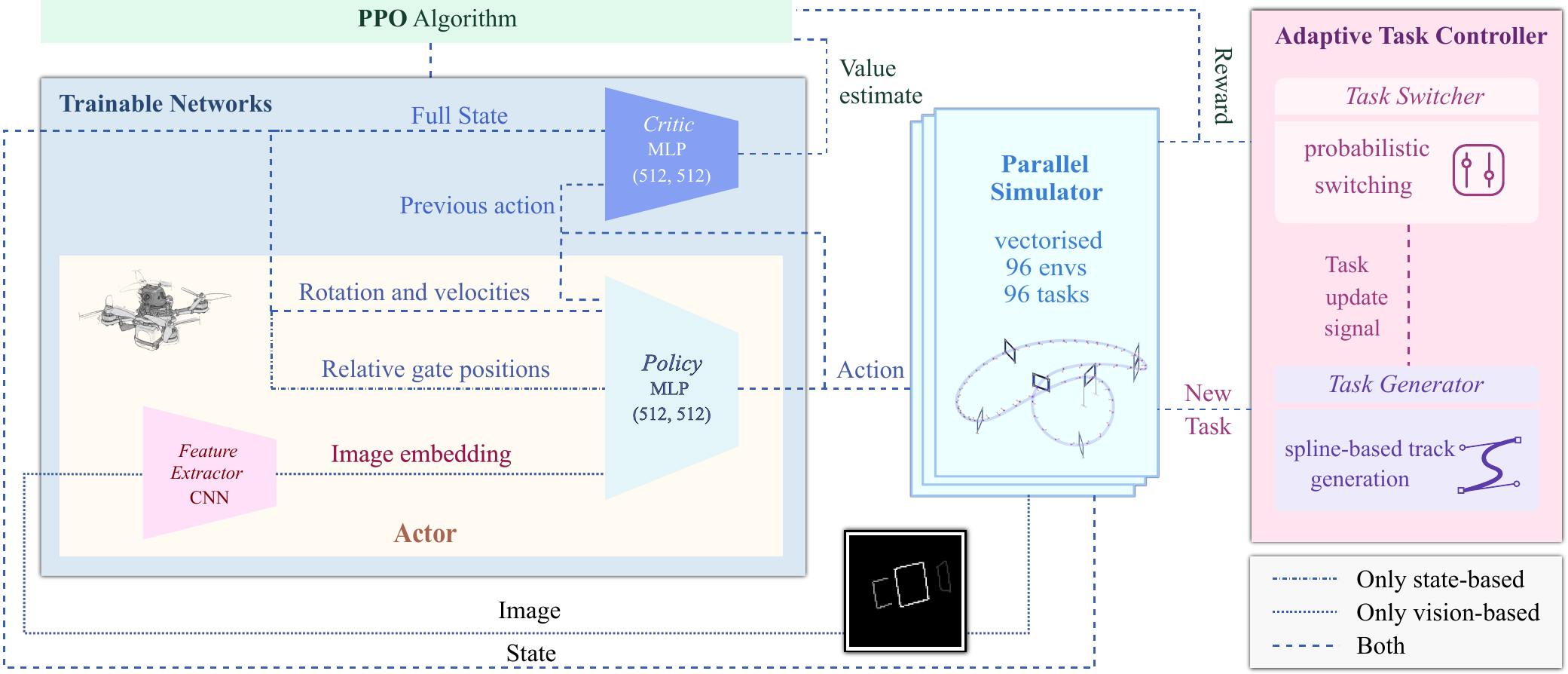}
    \vspace{0.1cm}
    \caption{Overview of the training pipeline and model architecture. For the state-based agent, the actor contains only the Policy network, which receives full state information and the previous action. For the vision-based agent, the Actor additionally contains a feature extractor CNN, and the Policy network receives the feature extractor's image embedding, partial state information, and the previous action. In both cases the set of training tasks are maintained by the Adaptive Task Controller, which assesses when tasks are no longer generating a strong learning signal to replace them with new tasks.}
    \label{fig:pipeline}
    \vspace{-0.5cm}
\end{figure*}

\section{Methodology}

\paragraph{Problem Formulation.}\label{subsec:problem_formulation}
We model the problem of ZSG in autonomous drone racing with an Underspecified Partially Observable Markov Decision Process (UPOMDP) inspired by \cite{parker-holder_evolving_2022}, defined by the tuple $\left(\mathcal{T}, \mathcal{O}, \phi, \mathcal{S}, \mathcal{A},\mathcal{P}, \mathcal{R}, \gamma\right)$ with discount factor $\gamma\in(0,1)$.
Similar to a conventional Markov Decision Process, $\mathcal{S}$ and $\mathcal{A}$ are the state and action spaces, respectively.
The task space $\mathcal{T}$ denotes the free parameters of the environment; in our case the gate positions and orientations (or racetrack layout).
The transition function mapping a given state and action to the next state is then $\mathcal{P}:\mathcal{S}\times\mathcal{A}\times\mathcal{T}\to\mathcal{S}$, and the reward function that computes the reward from a state is $\mathcal{R}:\mathcal{S}\times\mathcal{T}\to\mathbb{R}$.
We account for partial observability with observation space $\mathcal{O}$ and observation function $\phi:\mathcal{S}\to\mathcal{O}$.
We assume the environment parameter vector (which specifies the racetrack layout) does not change in a given UPOMDP. As different racetracks correspond to different UPOMDPs, we consider them different tasks, following \cite{vithayathil_varghese_survey_2020}. For generality, we consider a continuous task distribution $p(\mathcal{T})$, rather than a discrete set of tasks.
$\mathcal{M_\tau}$ denotes the UPOMDP induced by a parameter vector $\bm{\tau}\sim~p(\mathcal{T})$. We define $T_{\text{train}}\subset\mathcal{T}$ as the training set and $T_{\text{test}}\subset\mathcal{T}$ as the evaluation set.
Given this UPOMDP formulation, a policy space $\Pi$, and an initial state $s_0$, our objective becomes
$
    \pi^* = \max_{\pi \in \Pi} \mathbb{E}_{s_{0} \sim p(s_{0}), \bm{\tau} \sim p(\mathcal{T})} [R(\pi, \mathcal{M}_{\bm{\tau}})].
$
Hence, we seek to find the single policy $\pi^*$ that performs best in expectation across the distribution of possible tasks, not just any single task.
\paragraph{Policy Learning: From State to Vision.}\label{subsec:model}
In contrast to conventional PPO, we train on multiple different tasks in parallel, and update $T_\text{train}$ during training based on feedback from the training process as described in Section \ref{subsec:improving_generalisation}.
At each timestep, the simulated environment provides an observation, and the agent outputs an action $\bm{a} = [c, \bm{\omega^\text{ref}}]$, where $c\in\mathbb{R}$ is the collective thrust and $\bm{\omega^\text{ref}}\in\mathbb{R}^3$ the body-rate setpoints for roll, pitch, and yaw. 
To validate our proposed framework we consider observations from two modalities: state and vision.
Vision-based control is essential and particularly challenging, as autonomous drone racing is inherently partially observable and relies on visual cues to infer unseen track geometry.

Following \cite{wang2025environment}, the state observation is
$
\bm{o}^{\text{state}} = \bigl[ \bm{\tilde{R}}, \bm{v}, \bm{\omega}, \bm{a}_{\text{prev}}, \delta \bm{p}_1, \delta \bm{p}_2 \bigr],
$
where $\bm{\tilde{R}}\in\mathbb{R}^6$ encodes the first two columns of the rotation matrix $\bm{R}_{WB}$ \cite{zhou_continuity_2020}, $\bm{v}\in\mathbb{R}^3$ and $\bm{\omega}\in\mathbb{R}^3$ are the linear and angular velocities, and $\bm{a}_{\text{prev}}$ is the previous action. 
Gate-related information is provided by $\delta \bm{p}_1,\delta \bm{p}_2\in\mathbb{R}^{12}$, which encode the relative positions of the drone to the four corners of the upcoming gate and from the upcoming gate to the subsequent gate, respectively.

The vision observation $\bm{o}^{\text{image}}$ is an $84\times84$ image processed by a three-layer Convolutional Neural Network (CNN) encoder into a 128-dimensional embedding $\bm{o}^{\text{embed}}$.
While previous vision-based drone racing agents such as \cite{geles_demonstrating_2024} have used black and white images, we instead use 8-bit to encode gate ordering (necessary for determining gate order).
Specifically, we make the next gate full brightness, with subsequent gates decreasing in brightness (visible in Figure~\ref{fig:pipeline}).
For state-based training, both actor and critic receive $\bm{o}^{\text{state}}$.
For vision-based training, the critic again uses $\bm{o}^{\text{state}}$, while the actor receives $[\bm{o}^{\text{embed}}, \bm{\tilde{R}}, \bm{v}, \bm{\omega}]$. Inclusion of $\bm{\tilde{R}}$ ensures rotational stability in cases of partial gate visibility, and $\bm{v},\bm{\omega}$ provide explicit velocity information otherwise unobtainable from the embedding alone. 
The reward structure follows \cite{wang2025environment}, with minor modifications to the vision-based agent to encourage keeping gates within the camera's field of view.
Details can be seen in Appendix \ref{app:reward_structure}.

\begin{table}
    \centering
    \caption{Simulated and real performance of the generalist agent (Ours), Environment as Policy (EaP), and single-task (ST) policies. Note that for ST a dedicated agent was trained for each track, while for ours and EaP a single policy was deployed across all tracks. S2R Gap denotes the sim-to-real gap, measured as a percentage change in lap time between simulated and real-world deployment.}
    \begin{tabular}{rccccc}
    \toprule
        & \multicolumn{3}{c}{\raisebox{0.2ex}{Lap Time: Simulated (s)}} 
        & \multicolumn{2}{c}{\raisebox{0.7ex}{Lap Time: Real-World}} \\
        Track 
        & Ours. & EaP & ST 
        & Ours. (s) & S2R Gap (\%) \\
        \midrule
        \grayrow
        \it Figure8 & 2.940$\pm$0.4376 & 4.746$\pm$0.1952 & 2.546$\pm$0.2145 & 3.273$\pm$0.042 & 11.33 \\
       \it BigS    & 4.514$\pm$0.2539 & 7.513$\pm$0.4877 & 
       4.099$\pm$0.1991 & 4.633$\pm$0.076 & 2.636 \\
        \grayrow
        \it Kidney  & 3.202$\pm$0.2564 & 4.943$\pm$0.2962 & 2.340$\pm$0.1709 & 3.511$\pm$0.106 & 9.650 \\
        \it SplitS  & 5.140$\pm$0.4157 & crash & 4.700$\pm$0.1963 & 5.278$\pm$0.021 & 2.685 \\
        \bottomrule
    \end{tabular}
    \label{tab:combined_results}
    \vspace{-0.4cm}
\end{table}

\paragraph{Improving Generalization.}\label{subsec:improving_generalisation}
To generalize across a distribution of tasks, the agent must learn transferable skills rather than overfit to a single environment. 
In drone racing, this corresponds to training across multiple racetracks so the policy can handle arbitrarily complex, realistic gate layouts.
Prior approaches for multi-task RL, such as \cite{parker-holder_evolving_2022, xing2024multi, jiang_prioritized_2021}, are based on sequential curricula. 
These are, by their nature, susceptible to catastrophic forgetting, in which performance on earlier tasks degrades when new ones are introduced \cite{french_catastrophic_1999}.
We demonstrate the severity of this in Appendix \ref{app:forgetting}.

To combat this, we adopt a parallelized training paradigm in which rollouts are collected across multiple different racetracks simultaneously and used jointly for policy updates. 
Additionally, we propose new methods for assessing when a task in $T_\text{train}$ is no longer generating a strong learning signal and for generating new tasks to train on. 
We refer to these processes as \textit{Adaptive Task Switching} and \textit{Informed Task Generation} respectively, and assess their impact, along with the more conventional $L_2$ regularization method (Appendix \ref{app:l2reg}) on improving ZSG for autonomous drone racing.
We measure generalization using the quotient of success rate and lap time, averaged across all evaluation tasks, which we refer to as \emph{Performance-Weighted Success Score}, or $S_{pw}$. We use this metric as it captures both feasibility \emph{(can the agent finish a lap?)} and performance \emph{(how quickly?)}, which existing generalization evaluations in RL typically do not. Detailed discussion is available in Appendix \ref{app:generalisation}.

\begin{figure*}[t]
    \centering
    \includegraphics[width=1.0\linewidth]{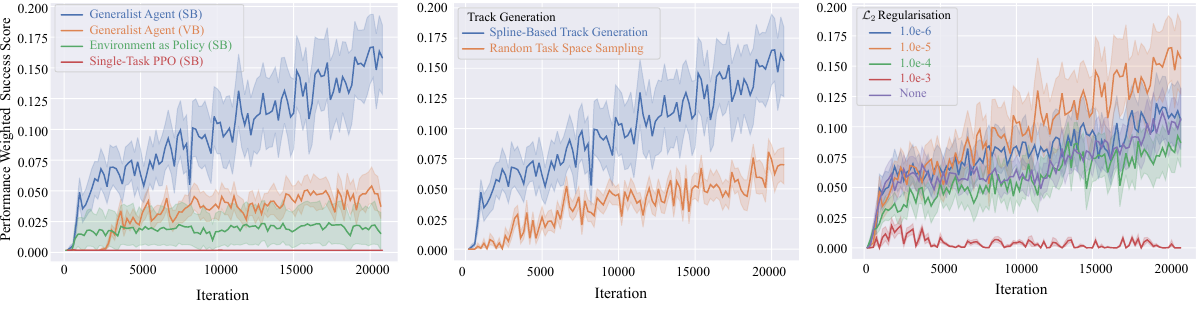}
    \caption{\textbf{Left:} Comparison of the generalization performance of this and other works. ``SB'' denotes models with state-based observation format, while ``VB'' denotes vision-based. Error bands indicate a 95\% confidence interval.
    \textbf{Middle:} Comparison of generalization during training when using different track generation methods.
    \textbf{Right:} Comparison of the impact of $L_2$ regularization on generalization during training.
    }
    \label{fig:reward_merged}
    \vspace{-1.5em}
\end{figure*}

\paragraph{Informed Task Generation.}
A straightforward method for producing new training tasks is uniform random sampling from the task space. 
However, this has several drawbacks. 
First, uniform sampling does not respect the underlying structure of $\mathcal{T}$: feasible drone racing tracks are not distributed uniformly across the high-dimensional task space. 
Second, as dimensionality grows, maintaining sufficient sample density requires exponentially more samples, causing sparsity of exploration. 
Finally, randomly sampled tracks may be either infeasible (violating physical constraints) or uninformative (providing little learning signal).

We therefore introduce an informed task generation approach that restricts sampling to a subspace of $\mathcal{T}$ defined by feasibility and diversity criteria. 
Specifically, tasks must (i) admit at least one smooth trajectory through all gates, (ii) satisfy physical and geometric constraints, and (iii) vary sufficiently in geometric properties to expose the agent to a wide range of flight dynamics. We achieve this by fitting a B-Spline to points in 3D space, then placing gates tangent to the curve at arc-length spaced intervals.
This process is formally defined, and constraints elucidated in Appendix \ref{app:gen_math}.

\paragraph{Adaptive Task Switching.}
Assessing when a task should be replaced requires measuring whether it continues to generate a useful learning signal.
Using raw reward thresholds such as in \cite{wang_paired_2019} is unsuitable in our setting, since reward scales differ between tracks.
Given the computational difficulty associated with computing time-optimal trajectories, it is infeasible to normalize these reward scales with a maximum reward a priori. 
Furthermore, task difficulty is non-stationary: as the policy improves, tasks within the training set transition from challenging to trivial.

Instead of absolute performance, we focus on performance improvement. 
If the expected return for a task ceases to change across recent updates, then that task is no longer providing gradients that yield meaningful policy updates. 
Since rollouts from multiple tasks are collected in parallel, global reward trends may obscure local convergence, necessitating task-specific analysis.

We model this as an \emph{online flatness detection problem}, where we seek to determine (per task) how fast the reward curve flattens, and propose two methods for detecting reward flatness: a statistical method using Spearman's rank correlation coefficient \cite{hollander_nonparametric_2013}, and a Kalman filter method using a Local Linear Trend model \cite{laine2019introduction}. 
Details are available in the Appendix \ref{app:switching}.

Both methods output a confidence score representing a likelihood that the reward has recently plateaued. At each training iteration, this confidence score is computed as a probability $p_{\text{switch}}$ for each task and is used as the likelihood of switching to a new one. This mechanism ensures that training time is allocated to tasks whose returns are actively changing, rather than to those that have saturated and are beginning to overfit.

\begin{wrapfigure}[15]{r}{0.6\textwidth}
    \centering
    \vspace{-12pt}
    \includegraphics[width=0.9\linewidth]{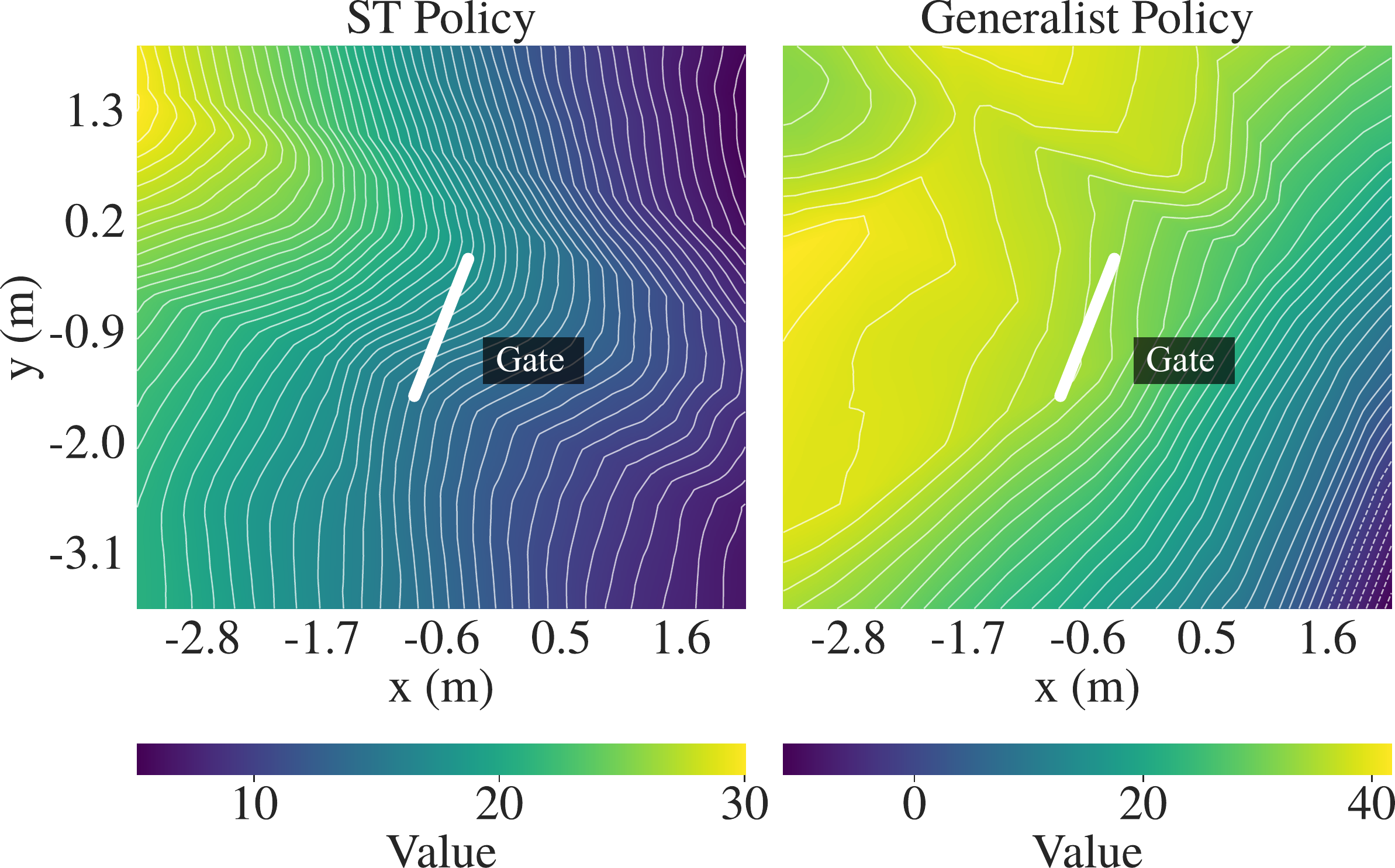}
    \caption{Value function as predicted by the critic plotted for a fixed observation across the XY plane for the first gate of the SplitS racetrack.}
    \label{fig:critic_heatmap}
\end{wrapfigure}

\section{Results and Discussion}\label{sec:discussion}

We compare our generalist agent (Section~\ref{subsec:improving_generalisation}) with Environment as Policy (EaP)~\cite{wang2025environment} and single-task (ST) PPO~\cite{kaufmann_champion-level_2023}. Policies are evaluated on unseen racetracks, including generated tracks and four human-designed core tracks (Figure8, BigS, Kidney, and SplitS; Figure~\ref{fig:catcheye}) drawn from \cite{wang2025environment}. We also deploy the state-based policy to these core tracks in the real world. Training, evaluation, and deployment details are given in Appendix \ref{app:train_eval_deploy}.

\begin{figure}[t]
    \centering
    \includegraphics[width=0.8\linewidth]{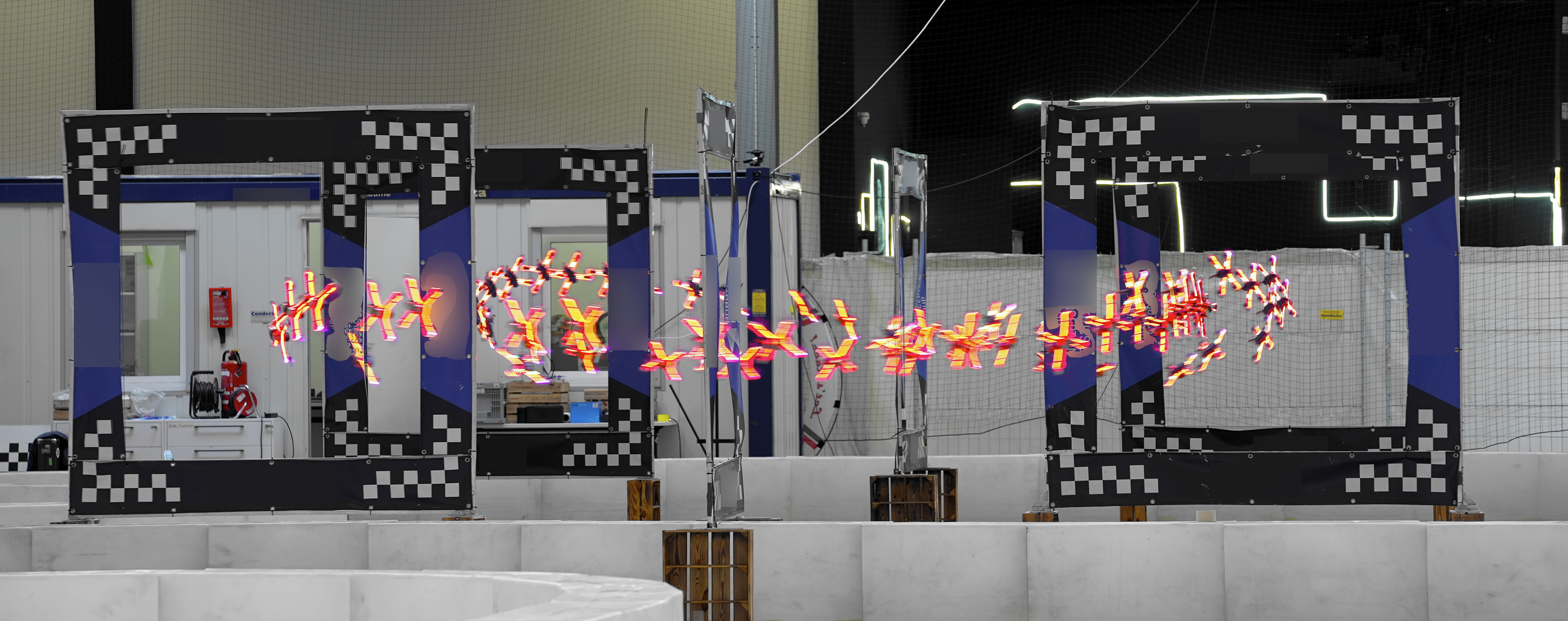}
    \vspace{-0.2cm}
    \caption{Real-world deployment of the generalist agent on the Figure8 racetrack.}
    \label{fig:realworld}
    \vspace{-0.45cm}
\end{figure}

\textbf{How effective is our approach?} 
The generalist agent is capable of high-speed racing on a broad range of unseen tracks. The state-based model reaches a maximum $S_{pw}$ of 0.1652, compared with 0.0222 for EaP and \SI{3.250e-4} for ST, corresponding to a $7.4\times$ improvement over the current state of the art (Figure~\ref{fig:reward_merged} Left). Even with EaP's thrust limited to 6.0N to improve its generalization, our method achieves much higher success rates (69.8\% state-based / 60.8\% vision-based vs.\ 2.4\% EaP / 0.0\% ST).
This generalization gain does not come at the expense of competitive speed: the state-based generalist is 37.73\% faster than EaP across the three core tracks completed by EaP, and only 14.52\% slower than per-track ST agents that do not generalize (Table~\ref{tab:combined_results}). It also transfers to hardware, completing all four core tracks with a 100\% success rate and a mean sim-to-real lap-time increase of only 6.575\% (Figure~\ref{fig:realworld}). Finally, the vision-based generalist reaches $S_{pw}=0.05219$, which is $2.3\times$ better than state-based EaP despite the harder observation format. To our knowledge, this is the first demonstration of ZSG for vision-based high-speed drone racing.

\textbf{What are the key contributions for improved ZSG?}
To identify which components affect ZSG, we disable parts of the state-based training pipeline. With $L_2$ regularization disabled, random task switching slightly reduces maximum $S_{pw}$ by 2.963\% relative to no switching ($S_{pw}=0.1004$), whereas adaptive switching improves it by 10.29\% with Spearman flatness detection and 10.27\% with the Kalman method. The two adaptive variants behave similarly, each switching tasks approximately 3000 times during training.

Informed task generation is also critical. Replacing our spline-based generator with uniform random sampling over gate positions and orientations reduces the best score from $S_{pw}=0.1652$ to $S_{pw}=0.08069$, a 2.05$\times$ drop (Figure~\ref{fig:reward_merged} Right). Weight decay improves ZSG when tuned, with the best result at \SI{1.0e-5}{}, but values of \SI{1.0e-4} or larger harm performance or destabilize training. For vision-based control, convergence requires a compact image embedding, a small set of state channels (rotation, velocity, angular rates), and the gate-brightness encoding described in Section~\ref{subsec:model}.

\begin{wrapfigure}[22]{r}{0.6\textwidth}
    \centering
    \vspace{-0.4cm}
    \includegraphics[width=\linewidth]{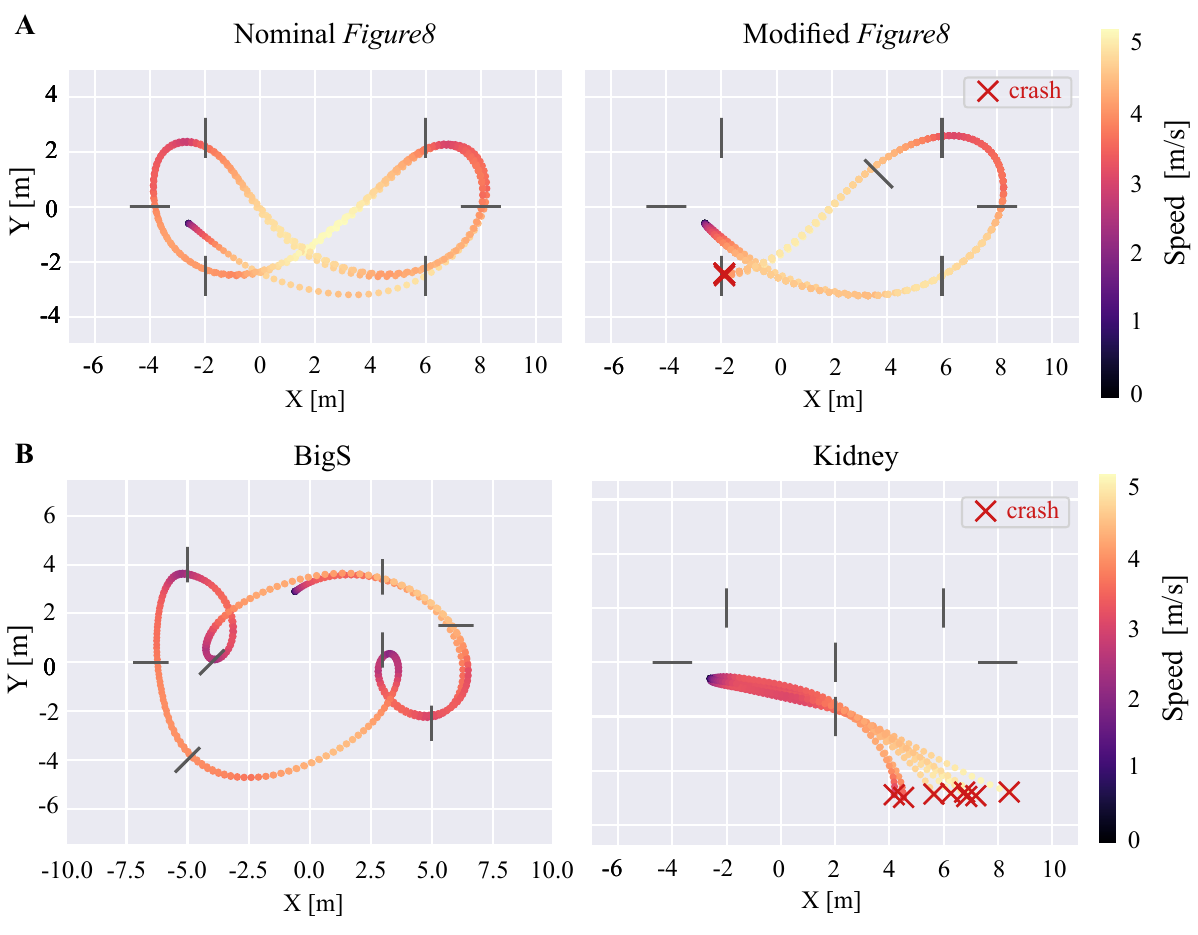}
    \caption{Policies trained on a single track effectively memorize a single trajectory.
    (\textbf{A}) Providing a small change to the observation (new gate along existing trajectory) induces a crash. 
    (\textbf{B}) The agent (incorrectly) turns right on an unseen track and crashes because the first turn during training track was rightwards.
    }
    \label{fig:traj_deviation}
\end{wrapfigure}

We do not find that recurrency improves generalization. An LSTM actor trained with state-based observations reaches a maximum $S_{pw}$ of 0.1431, 13.40\% lower than the MLP policy (see Appendix \ref{app:memory}). This reduction is mainly due to lower success rate; successful laps have comparable times. The recurrent policy also requires longer training and higher inference cost, suggesting that the state-based observation is already sufficiently Markovian for this task.

Our results thus identify several interacting mechanisms.
Adaptive task switching unlocks additional performance compared to not switching, with both the Spearman and the Kalman method proving effective (the Spearman method slightly more so). 
However, it is important to ensure switching only when the agent's performance on a task has plateaued, as random switching harmed ZSG ability.
Thus, the increased diversity of tasks provided by switching is helpful, but only when provided in a structured manner. 
A properly tuned weight decay (1.0$\times 10^{-5}$ in our experiments) similarly improves ZSG.
However, larger values destabilize learning, highlighting the sensitivity of regularization in on-policy RL.
Constraining the procedural generator to produce geometrically feasible and diverse tracks yields a 2.05$\times$ improvement over uniform sampling, suggesting that task quality matters as much as quantity.

\textbf{Why do Single-Task (ST) Policies fail?}
Unlike trajectory-tracking controllers~\cite{romero_model_2022,foehn_time-optimal_2021}, RL policies are trained without a reference trajectory. In practice, however, ST agents appear to collapse into implicit trajectory tracking: they discover one high-reward state-action sequence and overfit to it. We test this by adding a geometrically redundant gate to a trained Figure8 policy, which still destabilizes and crashes after clearing the inserted gate, despite requiring no deviation from the learned path (Figure~\ref{fig:traj_deviation}A). We also evaluate a BigS-trained policy on Kidney, where it turns the wrong way through the first gate (Figure~\ref{fig:traj_deviation}B) as the training track required this.

This suggests that ST policies encode the specific sequence of observations seen in training rather than learning a reusable strategy for unseen gate configurations. The critic value maps support this interpretation: the ST agent has a single high-value corridor, whereas the generalist has a broader high-value region through the gate (Figure~\ref{fig:critic_heatmap}). This broader value landscape is consistent with learning adaptable maneuvers instead of memorized trajectories, which also helps explain why ST policies are especially prone to catastrophic forgetting (see Appendix \ref{app:forgetting}).

\begin{figure}
    \centering
    \includegraphics[width=0.95\linewidth]{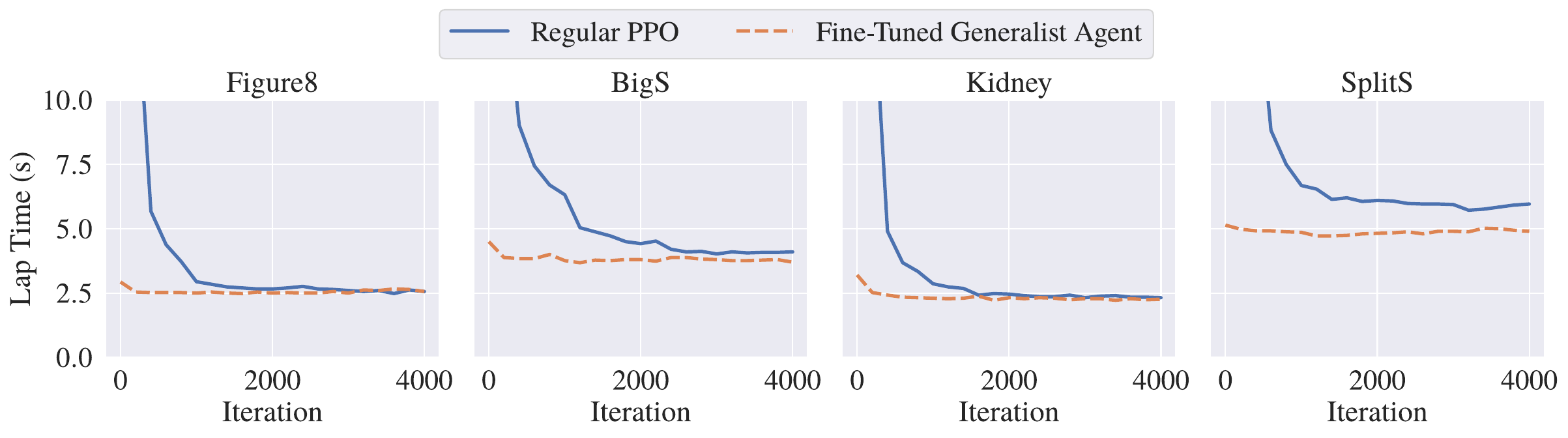}
    \caption{Fine tuning the generalist agent on a single track yields as good or better performance in a smaller number of iterations than training an ST agent from scratch.}
    \label{fig:distillation_training}
    \vspace{-2em}
\end{figure}

\textbf{What does this tell us about the trade-off between performance and generalization?}
Despite being widely acknowledged, poor ZSG in RL is not well understood~\cite{kirk_survey_2023}. Our results argue against a strict trade-off: the generalist policy is both 7.4$\times$ better at generalizing and 37.73\% faster than the next best method, while remaining only 14.52\% slower than ST agents that cannot generalize. This suggests that, at least in agile flight, architectural choices, data diversity, and structured task generation can recover much of the speed of specialized controllers while gaining large improvements in generalization. Moreover, after fine-tuning the generalist for \SI{1.0e+8} timesteps on one track with a reduced \SI{2.0e-4} learning rate, convergence requires only 23.19\% of the ST training iterations on average and yields lower lap times (Figure~\ref{fig:distillation_training}). Generalist pre-training is therefore also an efficient start point for task-specific adaptation and specialized high-performance policies.

\textbf{How does this compare to optimal control methods?}
Using the drone model parameters from MPCC~\cite{romero_model_2022,romero_time-optimal_2022} and MPCC++~\cite{krinner_mpcc_2024}, our generalist agent completes SplitS in 5.42s, compared with 5.67s for MPCC and 5.38s for MPCC++. This is comparable to state-of-the-art optimal control, but unlike MPCC-style methods, our policy does not require track-specific contouring-parameter tuning before deployment and therefore generalizes better.

\textbf{How does this approach apply to real-world contexts?}
The small sim-to-real gap~\cite{aljalbout2025reality, ren2026learning,pan2026learning} and vision-enabled results indicate practical promise: a generalist policy can reduce deployment effort in new venues and provide a strong fine-tuning starting point. Nevertheless, safety-critical deployment requires additional safeguards, including provably safe fallback controllers and more reliable perception under partial gate visibility, lighting variation, and motion blur. From a systems perspective, reducing the need for external motion-capture systems, for example, through stronger onboard perception or additional sensors such as event cameras, remains an important next step. We include real-world experiment videos in the supplementary material.

\textbf{What are the limitations of our framework?}
Several limitations remain. Extrapolation to out-of-distribution track families, dynamic gates, and novel obstacles is not yet characterized. The vision-based agent still requires partial state information for stability, which limits its autonomy compared with a purely image-based policy. Large-scale parallel training is also computationally demanding and currently depends on a simulator that can generate many diverse tracks quickly. Future work should therefore target purely vision-based policies, stronger safety mechanisms, broader out-of-distribution evaluation, and improved sample efficiency.

\vspace{-0.15cm}
\section{Conclusion}
\vspace{-0.15cm}
This work introduces and evaluates a training paradigm that enables zero-shot generalization in reinforcement learning for high-speed autonomous drone racing. Through diverse parallel training, informed task generation, adaptive task switching, and regularization, we train a single policy that robustly flies unseen racetracks in simulation and on real hardware, without task-specific fine-tuning. We additionally extend this to a vision-based observation format. To our knowledge, this is the first demonstration of zero-shot generalization in vision-based autonomous drone racing.
More broadly, our approach scales to high-dimensional task spaces and provides a practical framework for learning fast, generalizable policies without online adaptation. The adaptive task-switching mechanism offers a lightweight, automated alternative to manually designed curricula. Together, these results bridge the gap between generalization and performance in agile flight, advancing scalable and robust RL for real-world robotics.

\clearpage
\acknowledgments{This work was supported by the European Union’s Horizon Europe Research and Innovation Programme under grant agreement No. 101120732 (AUTOASSESS) and the European Research Council (ERC) under grant agreement No. 864042 (AGILEFLIGHT).}
\bibliography{bib/references, bib/references_OLD}  
\clearpage
\appendix
\section{Appendix}

\subsection{Reward Structure}\label{app:reward_structure}
At each timestep $t$, the reward is a weighted sum of components,
$$
r_t = r^{\text{prog}}_t + r^{\text{pass}}_t + r^{\text{crash}}_t + r^{\text{rate}}_t + r^{\text{cmd}}_t
$$

with coefficients $\lambda$ listed in Table~\ref{tab:weights}. Separate weightings are used for state-based (SB) and vision-based (VB) controllers.  

The progress reward reduces sparsity by providing incremental feedback:

\begin{equation}
r^{\text{prog}}_t =
\begin{cases}
\lambda_\text{prog}(d_{t-1}-d_t),
& \text{SB}, \\[4pt]
\lambda_\text{prog}\!\Big(
\max(d_{t-1}-d_t,0)\,\rho_t \\
\qquad\qquad
+ \min(d_{t-1}-d_t,0)
\Big),
& \text{VB}.
\end{cases}
\end{equation}

where $d_t$ is the Euclidean distance to the next gate center and $\rho_t\in[0,1]$, defined as the dot product between the normalized direction of the camera and the normalized normal to the gate plane, is a view-quality factor encouraging camera-gate alignment.

The gate-passing reward is triggered upon crossing the gate plane:
\[
r^{\text{pass}}_t =
\begin{cases}
  \lambda_\text{pass}\bigl(1 - e_t/w_g\bigr), & \text{SB} \\[4pt]
  \lambda_\text{pass} - \lambda_\text{error}\, e_t, & \text{VB},
\end{cases}
\]
where $e_t$ is the traversal error and $w_g$ half the gate width.  

Crashes incur a constant penalty $r^{\text{crash}}_t=\lambda_\text{crash}$.  

To prevent aggressive maneuvers detrimental to sim-to-real transfer, we penalize body-rate magnitudes:
\[
r^\text{rate}_t = \lambda_\text{rate}
   \left\|\tfrac{\omega_{t,xy}}{\omega^\text{max}_{xy}}\right\|^2 .
\]

Finally, to promote smooth and deployable control, the command penalty is defined as
\[
r^{\text{cmd}}_t = \sum_{i\in\{\text{thrust}, xy, z\}} 
   \lambda_{\text{diff},i} \Bigl( \lambda_\text{linear}|\Delta a_{t,i}| 
   + (\Delta a_{t,i})^2 \Bigr),
\]
where
\[
\Delta a_t = \frac{a_t - \tilde{a}_t}{a^{\max}},
\]
$a^{\max}$ is the maximum action magnitude, and $\tilde{a}_t$ is a first-order low-pass filtered action with a cutoff frequency of 6~Hz.  

\begin{table}[h]
    \centering
    \begin{tabular}{l|c|c|p{0.45\linewidth}}
        Term & SB & VB & Description \\
        \hline\hline
        $\lambda_\text{prog}$ & 1.0 & 0.8 & Progress towards next gate \\
        $\lambda_\text{pass}$ & 1.0 & 4.0 & Reward for gate traversal \\
        $\lambda_\text{error}$ & N/A & -1.0 & Penalty for inaccurate traversal \\
        $\lambda_\text{crash}$ & -4.0 & -4.0 & Collision penalty \\
        $\lambda_\text{rate}$ & -0.001 & -0.01 & Penalty on large body rates \\
        $\lambda_{\text{diff},\text{thrust}}$ & -0.0001 & -0.05 & Thrust smoothness penalty \\
        $\lambda_{\text{diff},xy}$ & -0.0002 & -0.01 & Roll/pitch smoothness penalty \\
        $\lambda_{\text{diff},z}$ & -0.0002 & -0.01 & Yaw smoothness penalty \\
        $\lambda_\text{linear}$ & 0.01 & 0.01 & Linear balancing factor \\
    \end{tabular}
    \caption{Reward weights for state-based (SB) and vision-based (VB) controllers.}
    \label{tab:weights}
\end{table}

\subsection{Spline-Based Track Generation}\label{app:gen_math}

\begin{algorithm}[H]
    \begin{algorithmic}[1]
    \Require Maximum attempts $M$, number of control points $N_c$, number of gates $N_g$, world bounds $B$, constraints $C$, bounded control point distribution $\mathcal{D}$
    \Ensure Valid track or failure
    \State $attempt \gets 0$
    \While{$attempt < M$}
        \State $p_i \sim \mathcal{D}(B,C) \quad \forall i \in \{1,\dots,N_c\}$
        \State $P \gets \{p_1, \dots, p_{N_c}\}$
        \State $S \gets \textsc{GenerateSpline}(P)$
        \State $(G, Q) \gets \textsc{PlaceGates}(S, N_g, C)$
        \State $valid \gets (G, Q \text{ satisfy all constraints in } C)$
        \If{$valid$}
            \State \Return $\textsc{BuildTrack}(G, Q, B)$
        \EndIf
        \State $attempt \gets attempt + 1$
    \EndWhile
    \State \Return \textsc{Failure}(``no valid track found'')
    \end{algorithmic}
    \caption{Informed Racetrack Generation}
    \label{alg:track_gen}
\end{algorithm}

Our method generates and represents racetracks using B-Splines. 
A set of $N_c$ three-dimensional control points $P = \{{\bm{p}_1},\dots,\bm{p}_{N}\}$ is sampled from a bounded distribution $\mathcal{D}(B, C)$.
A B-spline $S$ is fitted to these points, and gates are placed tangent to $S$ at arc-length-spaced intervals. 
The resulting spline defines a continuous reference trajectory through all gates. While this trajectory is not time-optimal, its smoothness guarantees the existence of at least one geometrically feasible flight path. 
Dynamic feasibility is approximated by enforcing curvature and spacing constraints $C$ during spline construction and gate placement.
By tuning spline parameters and control point distributions, we can generate novel racetracks that well approximate those seen in the real world, thereby aligning procedurally generated tasks with deployment conditions. Examples of human-designed and spline-based racetracks are available in Supplementary Materials.

The track generation process constructs a smooth, closed trajectory defined by a periodic spline that passes through a set of spatially constrained control points. Evenly spaced gates are then placed along this trajectory, each oriented according to the local tangent direction of the spline. The resulting path ensures both geometric feasibility and spatial smoothness suitable for drone racing tasks. This is described in Algorithm~\ref{alg:track_gen}.

\paragraph{Control Point Sampling}
A set of $N_c$ control points
\[
P = \{p_1, p_2, \ldots, p_{N_c}\}, \quad p_i \in \mathbb{R}^3
\]
is sampled uniformly within the bounded spatial region
\[
B = [x_{\min}, x_{\max}] \times [y_{\min}, y_{\max}] \times [z_{\min}, z_{\max}],
\]
subject to the following constraints:
\begin{align}
\text{(a) Horizontal spacing:} \quad & \|p_i^{xy} - p_j^{xy}\|_2 \geq d_{\min}, \quad \forall i \neq j, \\
\text{(b) Boundary padding:} \quad & x_{\min} + \delta_b \leq p_i^x \leq x_{\max} - \delta_b, \\
& y_{\min} + \delta_b \leq p_i^y \leq y_{\max} - \delta_b,
\end{align}
where $p_i^{xy} = [p_i^x, p_i^y]^\top$, $d_{\min}$ denotes the minimum horizontal spacing, and $\delta_b$ the required padding from the environment boundaries. The control points are re-centered within $B$ while preserving these constraints by translating all points by the average of their positions.

\paragraph{Periodic Spline Construction}
A closed periodic B-spline $\mathbf{s}(t): [0,1] \rightarrow \mathbb{R}^3$ is fitted through the control points:
\[
\mathbf{s}(t) = \text{B-spline}(P, s),
\]
where $s$ is a smoothing factor regulating curvature. The spline is defined by a tuple $(t, c, k)$ obtained via \texttt{scipy.interpolate.splprep}, enforcing periodicity:
\[
\mathbf{s}(0) = \mathbf{s}(1), \quad \mathbf{s}^{(n)}(0) = \mathbf{s}^{(n)}(1) \quad \forall n \leq k.
\]
This corresponds to the \textsc{GenerateSpline} function in Algorithm \ref{alg:track_gen}.

\paragraph{Gate Placement}
Let $L = \int_0^1 \|\mathbf{s}'(t)\| \, dt$ denote the total arc length of the spline. A set of $N_g$ gates is positioned at approximately equal arc-length intervals:
\[
g_i = \mathbf{s}(t_i), \quad \text{where} \quad \int_0^{t_i} \|\mathbf{s}'(t)\| \, dt = \frac{iL}{N_g}, \quad i = 0, \ldots, N_g - 1.
\]
Each gate is assigned an orientation quaternion $q_i$ derived from the tangent direction:
\[
\mathbf{t}_i = \frac{\mathbf{s}'(t_i)}{\|\mathbf{s}'(t_i)\|}, \quad
\theta_i = \arctan2(t_i^y, t_i^x), \quad
{q}_i = R_z(\theta_i),
\]
where $R_z(\theta_i)$ represents a rotation about the yaw axis aligning the gate with the local trajectory direction. Gate altitudes are clamped within
\[
z_{\min} + \tfrac{h_g}{2} \leq g_i^z \leq z_{\max} - \tfrac{h_g}{2}
\]
to ensure the full gate geometry remains inside the environment bounds. This corresponds to the \textsc{PlaceGates} function in Algorithm \ref{alg:track_gen}.

\paragraph{Start Pose}
The drone start pose is placed $2~\mathrm{m}$ behind the first gate along its negative forward axis:
\[
    x_{\text{start}} = g_0 - 2.0 \, \hat{f}_0,
\]
where $\hat{\mathbf{f}}_1$ is the forward direction derived from $q_0$. The starting attitude is identical to that of the first gate,
${q}_{\text{start}} = {q}_0$, so that the drone begins pointing towards the first gate; this is necessary particularly for the vision-based agent to be able to see the first gate in its image observation.

\paragraph{Constraint Validation}
A candidate track is accepted only if the following constraints are satisfied:
\begin{align*}
\text{Gate spacing:} \quad & \|{g}_i - {g}_j\|_2 \geq d_g^{\min}, \quad \forall i \neq j, \\
\text{Boundary constraint:} \quad &
x_{\min} + \delta_b \leq g_i^x \leq x_{\max} - \delta_b, \\
& y_{\min} + \delta_b \leq g_i^y \leq y_{\max} - \delta_b, \\
\text{Start position:} \quad &
x_{\min} \le x_{\text{start}}^x \le x_{\max}, \\
& y_{\min} \le x_{\text{start}}^y \le y_{\max}, \\
& z_{\min} \le x_{\text{start}}^z \le z_{\max}.
\end{align*}

These constraints form the set $C$ for Algorithm \ref{alg:track_gen}. If any of these constraints are violated, the generation process resamples control points and repeats until a valid configuration is found or the maximum number of attempts $M$ (as defined in Algorithm \ref{alg:track_gen}) is exceeded.

\newpage

\subsection{Training, Evaluation, and Deployment}\label{app:train_eval_deploy}

All policies are trained in 96 environments for $5\times10^8$ timesteps using the Flightmare simulator \cite{song_flightmare_2021} on an NVIDIA RTX 2080Ti. This corresponded to 2hrs of real-world time for the state-based model, 18hrs for the vision-based model, and 8hrs for the state-based LSTM described in Appendix \ref{app:memory}. For fairness, the ST approach is trained using the state-based observation format described Section \ref{subsec:problem_formulation}, meaning all three state-based methods have the same observation.

\begin{figure}[h]
    \centering
    \includegraphics[width=0.7\linewidth]{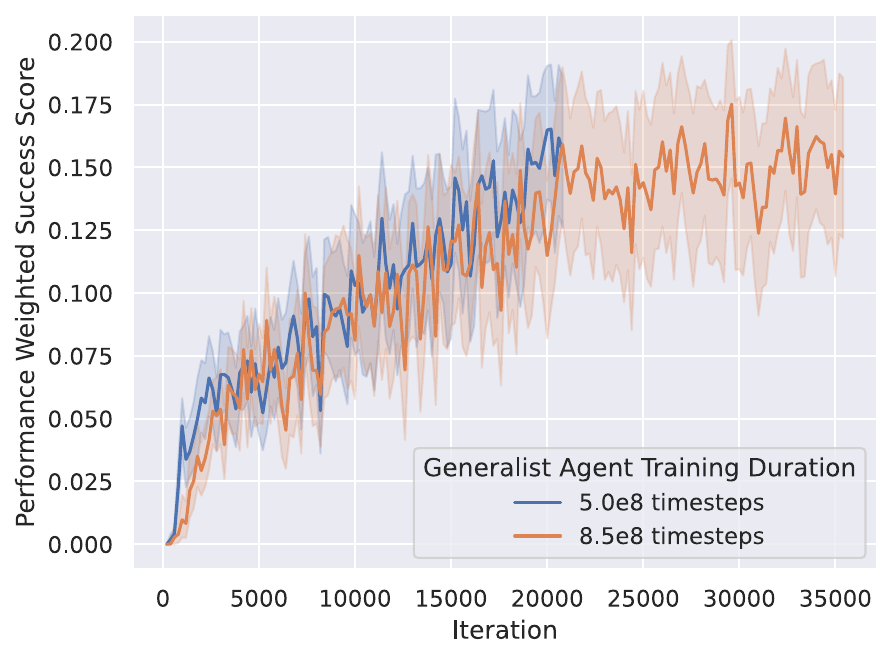}
    \caption{Training beyond $5\times10^8$ timesteps does not yield significant improvement in generalization ability}
    \label{fig:converge}
\end{figure}

For the generalist agent, experimentation showed no significant improvement in generalization (as measured by $S_{pw}$) when training longer than $5\times10^8$ timesteps, as visible in Figure \ref{fig:converge}. The EaP and ST baselines converge within approximately $2.5\times10^7$ timesteps (Figure \ref{fig:reward_merged}). Given this, we do not train beyond $5\times10^8$ timesteps when we are comparing methods.

For real-world deployment of the state-based model, a VICON motion capture system is used for state estimation. Inference on the control policy network is performed on a workstation and transmitted to the quadrotor, where the collective thrust and body rates are converted to motor commands by a low-level controller with Betaflight firmware.
Measuring the ability of a generalized drone racing agent requires evaluating its performance on a large number of different racetracks, all of which are unseen (i.e., not present in the set of training tasks). 
Specifically, we use a test set of 40 unseen racetracks: nine human-designed and 31 generated with the spline-based method described in Section \ref{subsec:improving_generalisation}. The tracks vary in number of gates from six to eight.
Each agent is deployed 25 times per track for 1000 timesteps in simulation, and for six laps on real racetracks. The initial position is varied randomly within $\pm 0.8$~m in the x and y direction and $\pm 0.6$~m in the z direction to ensure the policy's robustness. Scores are averaged over successful trials.

Complete track specifications, and examples of human-designed and spline-based racetracks are available in Supplementary materials.

\subsection{Adaptive Task Switching}\label{app:switching}

\begin{algorithm}[h]
\caption{Parallelized Proximal Policy Optimization with Task Switching}
\begin{algorithmic}[1]
\Require Task generator $\mathcal{G}$, task switching function $\text{Switch}: R_i \mapsto [0,1]$, $N$ environments, initial policy parameters $\theta$, initial value function parameters $\phi$, and learning rate $\eta$
\Ensure Optimised policy parameters $\theta^{*}$

\State Initialise task parameters $\{ \tau_i \}_{i=1}^N \sim \mathcal{G}$
\State Initialise environments $\mathcal{E}_i(\tau_i)$ for $i = 1 \dots N$

\For{each iteration}
    \State Collect trajectories $\lambda_i$ from $\pi_\theta$ in all $\mathcal{E}_i(\tau_i)$
    \For{$i = 1 \dots N$}
        \State $p_i \gets \text{Switch}(R_i)$
        \State With probability $p_i$: $\tau_i \sim \mathcal{G}, \; \mathcal{E}_i \gets \mathcal{E}_i(\tau_i)$
    \EndFor
    \State Estimate advantages $\hat{A}_t$ and returns $\hat{R}_t$ from $\{ \lambda_i \}$
    \State Form total PPO loss $L(\theta,\phi)$ with $\hat{A}_t$ and $\hat{R}_t$
    \State Update $(\theta,\phi) \gets (\theta,\phi) - \eta \nabla_{\theta,\phi} L(\theta,\phi)$
\EndFor
\end{algorithmic}
\label{alg:ppo_task}
\end{algorithm}

We integrate the adaptive task switching into the conventional PPO training method using the algorithm defined in Algorithm \ref{alg:ppo_task}. For the switching function $\text{Switch}: R_i \mapsto [0,1]$, we propose two methods, which are described below. For both methods, let $R_i^\tau$ denote the cumulative rollout reward on task $\tau$ at policy iteration $i$. 

\subsubsection{Spearman Switching Method}
For this method, we consider a sliding window of length $l$ which contains the last $l$ cumulative rollout rewards. In practice, we set $l=600$. 
To measure trend changes in reward during training, we compute Spearman's rank correlation coefficient $\rho$ \cite{hollander_nonparametric_2013} between these cumulative reward values and their corresponding timesteps for each task. Because the timesteps are monotonically increasing, this tells us that our reward trends upwards with $\rho > 0$, trends downwards with $\rho < 0$, and is flat with $\rho \approx 0$.
To map this statistic into a switching score, we define a smooth heuristic function:
$$
p_{\text{switch}}([R_{i-l}^\tau, \dots, R_i^\tau]) = \frac{1}{1+\alpha\rho^2}
$$
where $\alpha > 0$ controls sensitivity.
This score $p_{\text{switch}} \in (0,1]$ is a monotonic confidence-like measure where higher values correspond to greater certainty that the reward curve is flat. We set $\alpha=1000$ for all our trainings. Both positive and negative slopes reduce the score, since either indicates deviation from stagnation.

\subsubsection{Kalman Switching Method}
For this method, we use a Local Linear Trend model with a Kalman filter \cite{laine2019introduction} to directly compute the likelihood that the trend in the reward curve is flat.
The filter has state $x_i = [\ell_i,\ t_i]^\top$ (level, trend), and evolves as
\[
x_{i+1} =
\begin{pmatrix}
1 & \Delta t\\
0 & 1
\end{pmatrix} x_i + w_i,
\qquad
w_i \sim \mathcal{N}\!\left(0,\,
\mathrm{diag}(\sigma^2_{w,\ell},\sigma^2_{w,t})\right),
\]
with observation model
\[
R^\tau_i = [1\ \ 0]\,x_i + v_i,
\qquad
v_i \sim \mathcal{N}(0,\sigma^2_{\text{obs}}).
\]
where $\sigma_{w,\ell}$ and $\sigma_{w,t}$ are the process noise variance of the level and trend respectively, and $\sigma_\text{obs}$ is the observation noise variance.

After filtering $T$ observations, the posterior of the trend component is Gaussian,
\[
t_T \mid R^\tau_{1:T} \sim \mathcal{N}(\mu_t,\sigma_t^2),
\]
where $(\mu_t,\sigma_t^2)$ are obtained from the Kalman posterior mean and covariance.

The switching probability is defined as the posterior mass of the trend within a tolerance
$\varepsilon \ge 0$:
\[
p_{\text{switch}}
= \Pr\!\left(|t_T| \le \varepsilon \mid y_{1:T}\right)
= \Phi\!\left(\frac{\varepsilon-\mu_t}{\sigma_t}\right)
 - \Phi\!\left(\frac{-\varepsilon-\mu_t}{\sigma_t}\right),
\]
where $\Phi(\cdot)$ denotes the standard normal CDF, meaning that we are computing ``How likely is it that the underlying trend in the reward curve has a gradient of magnitude less than $\varepsilon$?''. During training we use $\varepsilon=\SI{1.0e-2}{}$.

\subsection{Measuring Generalization}\label{app:generalisation}
In supervised learning, generalization is typically assessed using held-out test data. In RL, however, there is no consensus on how to measure generalization~\cite{zhang_study_2018,kirk_survey_2023}, and existing metrics are often task-dependent. 
We focus on ZSG within a fixed task distribution, where evaluation tasks are drawn from the same distribution as training tasks but are unseen during training, i.e. $T_{\text{train}} \cap T_{\text{test}} = \emptyset$.

Existing metrics for generalization in RL are not suitable for the autonomous drone racing context.
Namely, they:
(i) Require a precomputed optimal solution, e.g. \cite{parker-holder_evolving_2022}, which is computationally infeasible with a large number of tracks,
(ii) Consider success as a binary outcome, e.g. \cite{cobbe_quantifying_2019, yu_meta-world_2021}, which does not account for how well the agent completed the task (in our case, how fast it completes the track), or
(iii) Use reward as a direct proxy for performance, e.g. \cite{espeholt_impala_2018}, which fails to consider that the reward scale varies between tasks.
Given these limitations, we introduce the Performance-Weighted Success Score, $
S_{pw}(\pi, T_{\text{test}}) =
\frac{1}{|T_{\text{test}}|}
\sum_{\bm{\tau} \in T_{\text{test}}}
\text{success}(\pi,\bm{\tau})\,\text{score}(\pi,\bm{\tau}).
$
Here, $\text{success}$ is a binary indicator of task completion, and $\text{score}$ is a positive scalar reflecting the quality of completion. This formulation is problem-agnostic; success and score can be defined according to the domain. For drone racing, we set
%
%
\[
\text{success}(\pi,\bm{\tau}) = 
\begin{cases} 
1 & \text{$\geq1$ lap completed and no crashes},\\
0 & \text{otherwise},
\end{cases}
\]
\[
\text{score}(\pi,\bm{\tau}) = 
\begin{cases}
1/\text{fastest lap time}(\pi,\bm{\tau}) & \text{if successful},\\
0 & \text{otherwise}.
\end{cases}
\]

While $S_{pw}$ is not an absolute measure across different test sets, it provides a consistent and interpretable basis for comparing the ZSG ability of agents on the same $T_{\text{test}}$. 

\subsection{L2 Regularization}\label{app:l2reg} 
$L_2$ regularization is a method that prevents any single weight from deviating too far from zero by penalizing the sum of the squares of the weights \cite{farebrother_generalization_2020}. It is not commonly used in RL as it can decrease performance on the training dataset; however, it is widely used in other fields as a method for improving generalization \cite{moradi_survey_2020}. Some work suggests that it can also increase ZSG in RL \cite{cobbe_quantifying_2019, farebrother_generalization_2020}.

\subsection{Memory}\label{app:memory}

We also train an agent using an Long Short-Term Memory (LSTM) network \cite{hochreiter_long_1997} for the actor with the state-based observation (Section \ref{subsec:model}), and then deploy this policy in simulation to determine if introducing recurrent memory unlocks additional performance. In this case, the critic remains an MLP.
The LSTM policy does not generalize as well, in that it has a maximum $S_{pw}$ of 0.1431, 13.40\% lower than the MLP policy.
Note that this score is lower due to a lower success rate; the LSTM achieves comparable lap times on the unseen test set.
In addition, the more complex network architecture required significantly longer training time, and has a larger computational load for inference during deployment. 
We theorize that the highly observable nature of the state-based control pipeline and the reasonably Markovian dynamics mean that this is not a context in which the benefits of recurrent architectures can be utilized.

\newpage

\subsection{Catastrophic Forgetting}\label{app:forgetting}

\begin{figure}
    \centering
    \includegraphics[width=0.6\linewidth]{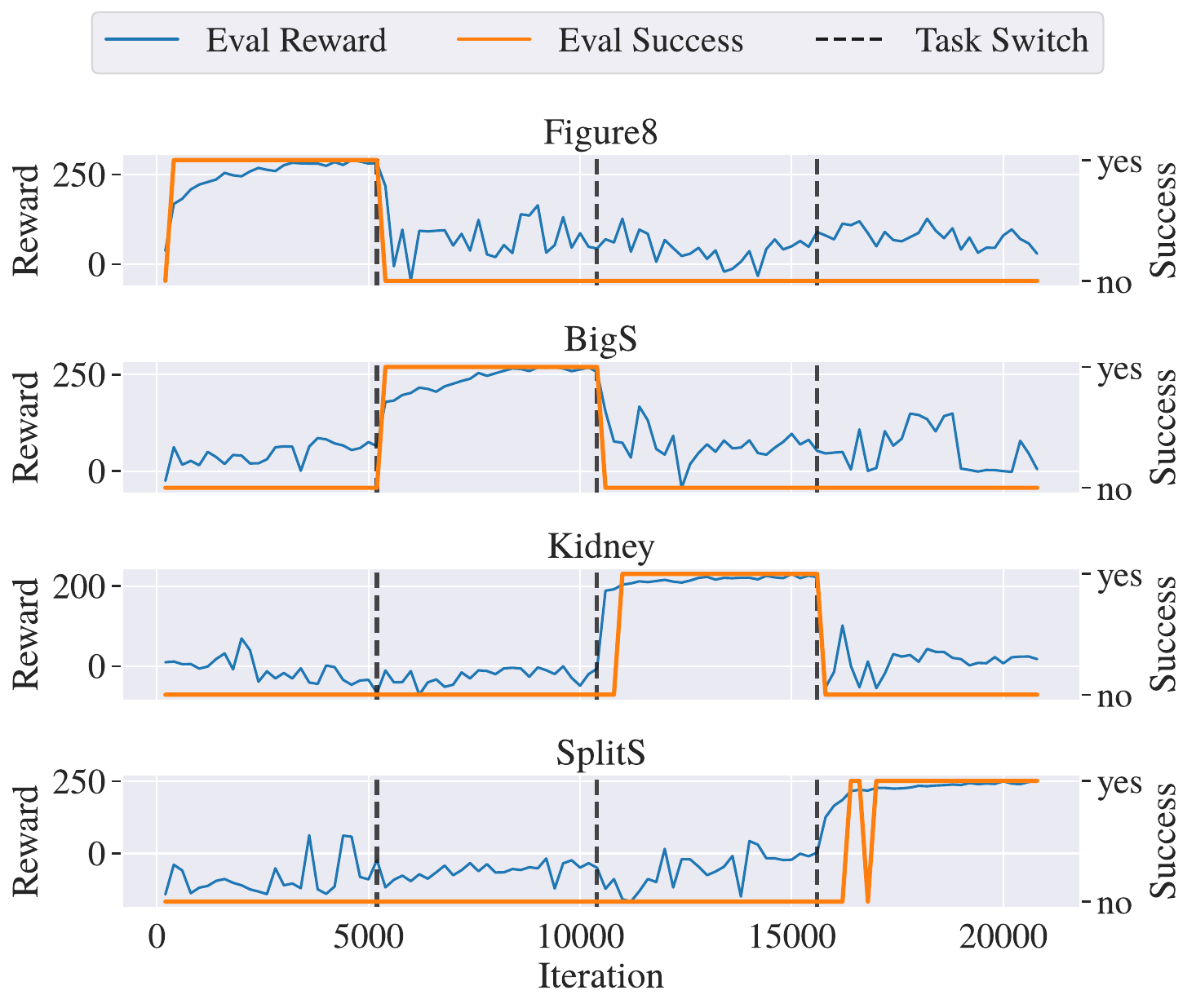}
    \caption{Demonstrating catastrophic forgetting. The training task switches every \SI{1.25e8} timesteps (5200 iterations), and the agent is only capable of completing the current task; performance collapses on previous racetracks when a new one is introduced.}
    \label{fig:forgetting}
\end{figure}

To demonstrate how acute catastrophic forgetting is in the context of agile flight, we train an ST racing agent on the four core tracks, each presented in isolation for \SI{1.25e8} timesteps.
As shown in Figure \ref{fig:forgetting}, the agent learns to complete the active track with high reward and success, but as soon as training switches, success on earlier tracks collapses to zero and is never recovered, despite previously high performance. We observe this with every task switch.

\end{document}